\documentclass[letterpaper, 10 pt, conference]{ieeeconf}  
\IEEEoverridecommandlockouts                              

\usepackage{cite}
\usepackage{amsmath,amssymb,amsfonts}
\usepackage{graphicx}
\usepackage{textcomp}
\usepackage{xcolor}
\usepackage{bm}
\usepackage{balance}
\usepackage{multirow}
\usepackage{makecell}
\usepackage{booktabs}
\usepackage{threeparttable}
\usepackage{pifont}
\usepackage{soul}
\usepackage[pagewise]{lineno}
\soulregister{\cite}7
\soulregister{\ref}7
\soulregister{\ding}7
\usepackage{float}

\usepackage{algorithm}      
\usepackage{algpseudocode}  
\usepackage{amsmath}        
\algnewcommand\algorithmicinput{\textbf{Input:}}
\algnewcommand\algorithmicoutput{\textbf{Output:}}
\algnewcommand\Input{\item[\algorithmicinput]}
\algnewcommand\Output{\item[\algorithmicoutput]}

\title{\LARGE \bf
Difference-Based Relational Learning for Zero-Shot Object-Goal Visual Navigation With Direct Sim-to-Real Transfer}
\author{Guolei Qi and Feitian Zhang*}

\begin{document}
\renewcommand{\hl}[1]{#1}
\maketitle

\thispagestyle{empty}
\pagestyle{empty}

\begin{abstract}
End-to-end deep reinforcement learning (DRL) for zero-shot object-goal visual navigation remains challenged by the sim-to-real gap, particularly variations in object appearance and restricted camera field-of-view (FoV). This letter proposes a Temporal Difference-Relational Network (T-DRN) for robust zero-shot sim-to-real transfer. T-DRN combines a Siamese difference-based feature extractor, which computes relational difference between the target and observed objects to produce domain-independent representations, with a dual-frame temporal buffer that preserves short-term object continuity under narrow FoV. Extensive experiments in AI2-THOR demonstrate that T-DRN improves zero-shot generalization in terms of  success rates over strong baselines. Furthermore, T-DRN is systematically validated on a physical wheeled robot, demonstrating robust performance under real sensing and actuation constraints and supporting the feasibility of direct sim-to-real transfer.
\end{abstract}

\section{Introduction}  
In object-goal visual navigation, an agent locates a target using only egocentric RGB observations. Because traditional map-based methods degrade in novel environments, practical deployment requires strong generalization to unseen objects and layouts. This challenge motivates zero-shot object-goal navigation (ZSON), in which the agent navigates toward object categories that are never encountered during training. Solving ZSON is critical for applications ranging from household service robots to logistics automation.

Recent end-to-end reinforcement learning approaches have shown promising results for ZSON in simulation \cite{SPNet, SAVN, SSNet}. However, their direct transfer to real-world scenarios remains an open challenge due to two fundamental issues. First, the sim-to-real gap introduces perceptual discrepancies---simulated objects differ significantly from their real-world counterparts in appearance, lighting, and texture. Second, real-world environments contain countless object categories, making it impractical to rely on explicit knowledge graphs or predefined object relationships.

Human cognition offers an elegant solution to this generalization challenge through \textit{difference-based learning}. Consider how a child learns to recognize a {zebra}: by comparing it with familiar animals, the child identifies discriminative features---the zebra shares a horse-like morphology but is distinguished by its {black-and-white stripes}; compared to a cow's irregular patches, the zebra's {regular, linear striping pattern} becomes its defining characteristic. This example suggests that comparing a novel object with familiar ones can reveal discriminative features that support robust recognition without explicit supervision.

Inspired by this observation, we propose the \textbf{Temporal Difference-Relational Network (T-DRN)} for zero-shot object-goal visual navigation. T-DRN explicitly models relational differences between the target object and all objects in view. Rather than relying on absolute representations, the network learns \textit{what makes an object distinct}, enabling recognition of unseen objects through their attributes relative to familiar visual elements.

Beyond object generalization, we identify another critical yet often overlooked aspect of the sim-to-real gap: the discrepancy in camera field-of-view (FoV). Standard simulation platform (e.g., AI2-THOR embodied AI environment \cite{SAVN, SSNet,tdanet}) typically assume wide-angle observations, which contrast with the restricted sensors available on physical platforms such as Turtlebot4, a wheeled robot adopted in this work. For end-to-end models that rely solely on RGB input, this sensing mismatch inherently causes severe performance degradation during direct sim-to-real transfer. To mitigate this issue, we introduce a {dual-frame temporal memory} mechanism that adapts the policy to narrow and variable FoVs without modifying the offline training data.

The main contributions of this letter are threefold.
\begin{itemize}
    \item We propose T-DRN, an object-centric relational difference architecture that models relational disparity between the target and all observed objects, improving  zero-shot generalization in object-goal navigation.
    \item We introduce a dual-frame temporal memory that addresses the FoV discrepancy between simulation and reality through lightweight short-term temporal buffering during direct policy transfer.
    \item We validate the proposed method through extensive experiments both in simulation and on a physical robot, where T-DRN achieves the strongest overall success rates among the compared methods and demonstrates practical direct sim-to-real transferability without fine-tuning.
\end{itemize}

\section{Related Work}

\subsection{Object-Goal Visual Navigation}
Object-goal visual navigation requires an agent to search for a target instance given only visual observations \cite{Zhu2017}. Many end-to-end deep reinforcement learning (DRL) models have been designed to establish navigation strategies that map observations directly to actions. Early studies focused on learning implicit representations of the observation before inputting it into the navigation policy \cite{SAVN}. For example, Wortsman \textit{et al.} \cite{SAVN} introduced self-adaptive visual navigation using meta-learning to adapt to test environments without explicit supervision. Other studies exploit object relationships or semantic contexts for more robust navigation policies \cite{mjol, Du2020}. More recently, heterogeneous scene representation frameworks have demonstrated that leveraging multi-level semantic relationships significantly enhances navigation efficiency in complex indoor scenes \cite{sociallyaware2024}. Specifically, MJOLNIR \cite{mjol} integrated hierarchical relationships among objects into a DRL model, achieving remarkable performance using object detection outputs. However, the static nature of their knowledge graphs often limits generalizability across novel scenes. While Transformer-based models like OMT \cite{Fukushima2022} and NavTr\cite{navtr2024} learn relationships dynamically, their high computational cost poses a barrier for real-time deployment on resource-constrained mobile robots. 
{In contrast, T-DRN adopts a lightweight object-centric design that processes detected objects individually. By focusing on fine-grained relational cues rather than global map building or computationally heavy transformers, T-DRN maintains high efficiency while preserving rich scene information.}

\subsection{Zero-Shot Object-Goal Navigation (ZSON)}
ZSON addresses the challenge of locating objects from categories that are not encountered during training \cite{SPNet}. While modular designs such as CoW \cite{Cow2022} and VoroNav \cite{VoroNav2024} leverage large multimodal models and depth images to generate global maps, this letter focuses on RGB-only DRL agents. Several strategies have been proposed to improve ZSON in end-to-end settings. Zhu \textit{et al.} \cite{Zhu2017} proposed a Siamese architecture for image-goal navigation across different scenes \cite{Siamese}. Khandelwal \textit{et al.} \cite{ZSON} utilized the zero-shot capability of CLIP \cite{Radford2021} to generate semantic embeddings of the goal. Recently, AKGVP \cite{AKGVP} leveraged the image-text matching ability of CLIP to align knowledge graphs with visual perception. Furthermore, the latest advances have explored the integration of large language models (LLMs) and semantic reasoning to achieve open-vocabulary zero-shot navigation \cite{osmag2026}. TDANet \cite{tdanet} introduced a target-attention module to selectively model the semantic and spatial relationships between the target and observed objects. However, attention-based methods often prioritize a small subset of relevant objects, potentially neglecting crucial contextual information.
{Building upon these concepts, T-DRN adopts a Siamese difference-based mechanism to model the relational disparity between the target and all observed objects. Relative to elective-attention schemes such as TDANet, the key distinction is that T-DRN aggregates target-object differences over the full set of detected objects and complements them with an explicit temporal buffer, thereby preserving broader scene context in zero-shot settings.}

\subsection{Real-World Deployment and Sim-to-Real Transfer}
While the aforementioned ZSON methods have demonstrated remarkable success in simulated environments, their transition to physical robots is frequently bottlenecked by the sim-to-real gap \cite{kadian2020sim2real}. This gap encompasses not only visual domain shifts but also severe geometric discrepancies. Although techniques like domain randomization can partially mitigate lighting and texture variations, the geometric mismatch caused by restricted camera FoV remains a critical yet under-explored challenge. Standard simulated environments (e.g., AI2-THOR \cite{ai2thor}) typically provide ideal, wide-angle panoramic observations, whereas physical robots are inherently constrained by much narrower sensing hardware. 
For existing single-frame end-to-end models, this hardware-induced sensory limitation amplifies partial observability, leading to severe perceptual aliasing and frequent target loss during agent rotation \cite{fang2019smt}. Addressing such hardware-constrained uncertainties and environmental changes is a prominent focus in recent mechatronics and robotics research, where robust image-based tracking \cite{feng2022tmech} and memory-enhanced architectures \cite{bioslam2023tro} are actively employed to compensate for sensing limitations on mobile platforms. Although some approaches attempt to build explicit semantic maps to overcome narrow FoVs \cite{chaplot2020semexp}, these methods are often computationally heavy for resource-constrained platforms. 
{To bridge this sim-to-real gap with a lightweight architecture, our work extends the difference-based framework with a short-term temporal memory mechanism. By explicitly incorporating a dual-frame temporal buffer, our model  ensures stable navigation toward unseen targets even under the restricted sensing capabilities of real-world platforms.}
\begin{figure*}[t]  
    \centering
    \includegraphics[width=\textwidth]{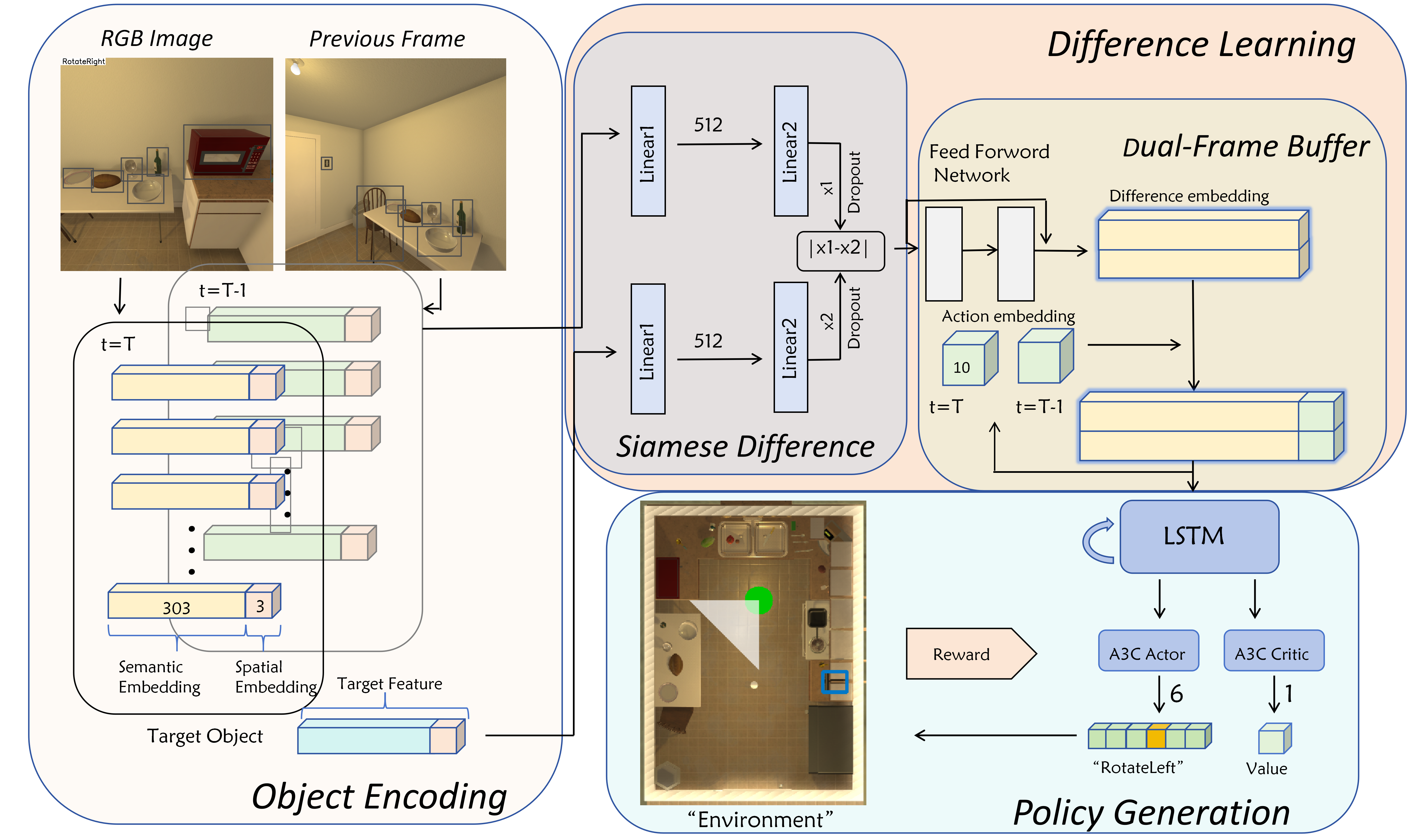}  
    \caption{Overview of the proposed T-DRN architecture. Dual-frame observations and the target object are  encoded into semantic and spatial embeddings. A Siamese difference module computes the relational features $|x_1 - x_2|$ between the target and observed objects. These features are fused with past actions and processed by an LSTM for temporal reasoning. The actor-critic policy then generates navigation actions.}
    \label{fig:model}
\end{figure*}

\section{Problem Formulation}

The ZSON task requires an agent to locate and navigate toward a specified target object $t$, which belongs to a predefined class set $T$. Unlike in simulation-only settings, our agent is now deployed on a physical TurtleBot4 robot, operating in real indoor environments with real-time sensory inputs and actuation uncertainties. The agent exclusively uses egocentric RGB observations without any prior access to a global map or environment layout.

This task becomes especially challenging in the {zero-shot visual navigation} setting, where the policy is trained on objects from a \textit{seen} class set $S$, but evaluated on locating objects from an \textit{unseen} class set $U$ ({i.e.}, $S \cap U = \emptyset$).
At the beginning of each episode, the initial positions of the agent and the target object are randomized within the test environment. The agent selects an action $a$ by executing a policy $\pi$ that receives the current RGB observation $I$ and the word embedding of the target object $w_t$ as inputs. Formally, $a \sim \pi_{\theta}(I, w_t)$, where $\theta$ represents the learnable parameters of the policy network.

The action space $\mathcal{A}$ is adapted to the TurtleBot4 platform and includes the following discrete actions: \texttt{MoveAhead}, moves the robot forward by approximately 0.25 meters; \texttt{RotateLeft} and \texttt{RotateRight}, rotates the robot by $45^\circ$ counter-clockwise or clockwise, respectively; \texttt{LookUp} and \texttt{LookDown}, adjusts the camera tilt by $30^\circ$; \texttt{Done}, signals the end of the episode when the agent believes it has found the target.

An episode is considered successful if the agent triggers the \texttt{Done} action while the target object is both \texttt{visible} in the current RGB frame and physically located within a 1.5-meter radius from the robot.

\section{Method}

The proposed T-DRN addresses the ZSON challenge by explicitly modeling the relational difference between the target and environmental observations. The architecture is designed to be lightweight and robust against geometric discrepancies encountered during sim-to-real transfer.

\subsection{Difference-Based Relational Learning}
For each detected object $i$ in the current observation, we construct a structured feature vector $V_i = [p_i, s_i] \in \mathbb{R}^{303}$. Here, $p_i = [x_i, y_i, a_i]$ represents the normalized spatial properties, including center coordinates and bounding box area, while $s_i \in \mathbb{R}^{300}$ is the semantic embedding obtained from a pre-trained GloVe model. The target object $T$ is represented by a similar 303-dimensional vector.

The feature extractor, implemented as a Siamese-style architecture, performs a point-to-point comparison between the target object $V_T$ and each observed object $V_i$. The process is formulated as follows:
\begin{equation}
    X_{ri} = \sigma(W_v V_i + b_v), \quad \text{and} \quad X_t = \sigma(W_v V_T + b_v),
\end{equation}
where $\sigma$ denotes the ReLU activation function. To explicitly capture the relationship between the observation and the goal, the model computes the absolute element-wise difference:
\begin{equation}
    d_i = | W_L X_{ri} - W_L X_t |.
\end{equation}
A residual transformation followed by Layer Normalization (LN) is applied to obtain the individual difference vector $V_{di}$:
\begin{equation}
    V_{di} = \text{LN}(\text{FFN}(d_i) + d_i).
\end{equation}

To handle a variable number of objects $n$ in the field of view, all individual vectors are aggregated using mean-pooling to produce a scene-level representation $X_t = \frac{1}{n} \sum_{i=1}^n V_{di}$.

\subsection{Dual-Frame Temporal Buffer}
To ensure navigation stability under restricted sensing conditions, we introduce a dual-frame temporal memory mechanism that utilizes temporal information.

\textbf{Temporal Buffering.} The model maintains a memory buffer of size $M=2$ to store the scene-level representations from the current and previous timesteps: $\mathcal{B} = \{X_{t-1}, X_t\}$. In scenarios where a restricted FoV leads to temporary object loss, the buffer provides the necessary historical context.

\textbf{Sequential Conditioning.} The temporal sequence from the buffer is concatenated with the embedded previous action probabilities $a_{t-1}$ to form the final input:
\begin{equation}
E_t = { [X_{\tau}, \sigma(W_a a_{\tau-1} + b_a)] }_{\tau=t-1}^{t}.
\end{equation}
This combined feature $E_t \in \mathbb{R}^{2 \times 522}$ is then processed by a Long Short-Term Memory (LSTM) network. The LSTM's hidden state $h_t$ is utilized by two separate linear heads to predict the navigation policy $\pi(a_t | h_t)$ (Actor) and the state-value $V(h_t)$ (Critic) within an Asynchronous Advantage Actor-Critic (A3C) framework.

\subsection{Reward Design}

The reward function $R$ is designed to guide the training of T-DRN by encouraging efficient navigation and learning object relationships in the environment. The reward balances task completion and exploration and is defined as:
\begin{equation}
R = \left \{
\begin{array}{rcl}
R_p  &      &  \text{if }p \text{ is } \texttt{visible}\\
R_t  &      & \text{if }t \text{ is } \texttt{visible} \text{ when } \texttt{Done}\\
R_t+R_p  &      & \text{if }t \text{\&} p\text{ are } \texttt{visible} \text{ when } \texttt{Done} \\
-0.01 &      & \text{otherwise}
\end{array} \right.
\label{eq_reward}
\end{equation}

Here, $R_t$ represents the reward for successfully locating the target object $t$, while $R_p$ is a partial reward based on the relationship  between the target object $t$ and its parent object $p$, which is defined in \cite{mjol} as the prior knowledge. The parent object $p$ refers to a larger object spatially or semantically associated with $t$, such as a table or refrigerator.  

When the agent performs the \texttt{Done} action and the target object $t$ is \texttt{visible}, it receives a reward of $R_t = 5$. Additionally, the partial reward $R_p$ encourages the agent to consider object relationships and is computed as: 
\begin{equation}
R_p = R_t \cdot {\rm Pr}(t \mid p) \cdot k,
\end{equation}
\noindent where $k = 0.1$ is a scaling factor, and ${\rm Pr}(t \mid p)$ represents the conditional probability of locating $t$ near $p$. This probability is calculated based on the relative spatial distances between $t$ and all parent objects $p$ in the scene \cite{mjol}. The agent earns $R_p$ when a parent object $p$ becomes \texttt{visible} for the first time. To encourage exploration, repeated observations of the same parent object do not yield additional rewards.  

Finally, a small penalty of $-0.01$ is applied at every step to promote efficient navigation and encourage shorter paths. By combining target rewards, partial rewards, and step penalties, the reward function ensures task success while fostering the agent's ability to understand object relationships, improving its generalization in zero-shot scenarios.

\section{Experiment} 
\subsection{Experimental Setup}
We employed the AI2-THOR virtual environment for training and testing. This environment comprises four categories of scenes: kitchen, living room, bedroom, and bathroom. Each scene type contains 30 rooms, with 20 designated as training scenes and 10 as testing scenes. Notably, the test set rooms are unseen during training. A commonly used set of 22 object classes is selected as the target set for navigation tasks.The agent is trained over 6,000,000 episodes using offline data generated from AI2-THOR v1.0.1, with a learning rate of 0.0001. During testing, 250 episodes per room type are evaluated.The performance of the agent is assessed using two widely adopted metrics in visual navigation tasks: success rate (SR) and success weighted by path length (SPL)~\cite{Li, metric}. The SR metric is computed as $\text{SR} = \frac{1}{N} \sum_{i=1}^{N} S_i$ where $N$ is the total number of episodes, and $S_i$ is a binary success indicator for the $i$-th episode. The SPL metric is calculated as $\text{SPL} = \frac{1}{N} \sum_{i=1}^{N} S_i \frac{L_i}{\max(L_i, e_i)}$ where $e_i$ denotes the agent's actual path length during the $i$-th episode, and $L_i$ is the optimal path length from the agent's initial position to the target object.

\subsection{Baseline Methods} 

We compare T-DRN against representative baselines models spanning random exploration, early target-driven DRL, graph-based reasoning, and recent zero-shot object-goal navigation methods.
\textbf{Random} selects its actions by sampling from a uniform probability distribution. \textbf{Zhu et al.} \cite{init1} takes as input the concatenation of ResNet features extracted from the current RGB image and the GloVe embedding of the target object.    
\textbf{MJOLNIR} \cite{mjol} introduces a novel context vector within a graph convolutional neural network to capture hierarchical object relationships for more effective navigation. 
\textbf{SSNet} \cite{SSNet} employs object detection outputs and cosine similarity of word embeddings as inputs to mitigate class dependency.  
\textbf{TDANet} \cite{tdanet} proposes a target-directed attention network that focuses on objects most relevant to the target within monocular visual observations, enhancing target-driven navigation performance.  

\subsection{Simulation Results} 

We evaluated the trained agents on two subsets of test episodes. The first includes all feasible episodes with optimal path length $L \geq 1$, while the second focuses on more challenging episodes with $L \geq 5$. This split makes it possible to assess both overall performance and behavior on longer-horizon navigation tasks.
\begin{table}[htbp]
\centering
\caption{Performance comparison on the 18/4 unseen-class split.}
\renewcommand{\arraystretch}{1.2}
\begin{tabular}{@{}lcccc@{}}
\toprule
\multirow{2}{*}{Model} & \multicolumn{2}{c}{$L\geq1$} & \multicolumn{2}{c}{$L\geq5$} \\
\cmidrule(r){2-3} \cmidrule(r){4-5}
& SR (\%) & SPL (\%) & SR (\%) & SPL (\%) \\
\midrule
Random & 10.8 & 2.1 & 0.9 & 0.3 \\
Zhu et al. \cite{init1} & 16.9 & 8.7 & 5.3 & 3.1 \\
MJOLNIR \cite{mjol} & 20.7 & 7.1 & 10.6 & 4.5 \\
SSNet \cite{SSNet} & 28.6 & 9.0 & 12.5 & 5.6 \\
TDANet \cite{tdanet} & 62.5 & 25.3 & 47.7 & 23.8 \\
\textbf{T-DRN (ours)} & \textbf{71.9} & 25.3 & \textbf{57.2} & \textbf{25.4} \\
\bottomrule
\end{tabular}
\label{tab:zs_18_4}
\end{table}

\begin{table}[htbp]
\centering
\caption{Performance comparison on the 14/8 unseen-class split.}
\renewcommand{\arraystretch}{1.2}
\begin{tabular}{@{}lcccc@{}}
\toprule
\multirow{2}{*}{Model} & \multicolumn{2}{c}{$L\geq1$} & \multicolumn{2}{c}{$L\geq5$} \\
\cmidrule(r){2-3} \cmidrule(r){4-5}
& SR (\%) & SPL (\%) & SR (\%) & SPL (\%) \\
\midrule
Random & 8.2 & 3.5 & 0.5 & 0.1 \\
Zhu et al. \cite{init1} & 14.6 & 4.9 & 4.9 & 2.8 \\
MJOLNIR \cite{mjol} & 12.3 & 5.1 & 6.0 & 3.6 \\
SSNet \cite{SSNet} & 21.5 & 7.0 & 13.0 & 6.7 \\
TDANet \cite{tdanet} & 53.4 & 20.4 & 41.2 & 20.7 \\
\textbf{T-DRN (ours)} & \textbf{62.7} & {20.2} & \textbf{42.7} & \textbf{21.3} \\
\bottomrule
\end{tabular}
\label{tab:zs_14_8}
\end{table}

Tables~\ref{tab:zs_18_4} and \ref{tab:zs_14_8} present the evaluation results of the comparison experiments on unseen object-goal navigation. T-DRN achieves the highest SR on both the 18/4 and 14/8 unseen-class splits. Relative to TDANet, which is the strongest competing baseline in these experiments, T-DRN improves SR by 8.2\% and 9.5\% on the 18/4 split for the $L \geq 1$ and $L \geq 5$ settings, respectively. On the 14/8 split, the improvements are 9.3\% and 1.5\%.

We hypothesize that the improved performance of T-DRN over TDANet is attributed to two design choices. First, TDANet relies on target-directed attention, which emphasizes a small subset of highly relevant objects. Although this improves action efficiency, it may neglect other informative objects that provide scene contexts and spatial cues, thereby limiting generalization. Second, T-DRN computes feature difference between the target and each observed object through a Siamese network and aggregates these difference representations across all detected objects. This design yields a more complete scene representation, reduces over-reliance on a few salient regions, and improves robustness in unseen environments.

\begin{figure}
    \centering
    \includegraphics[width=1.0\linewidth]{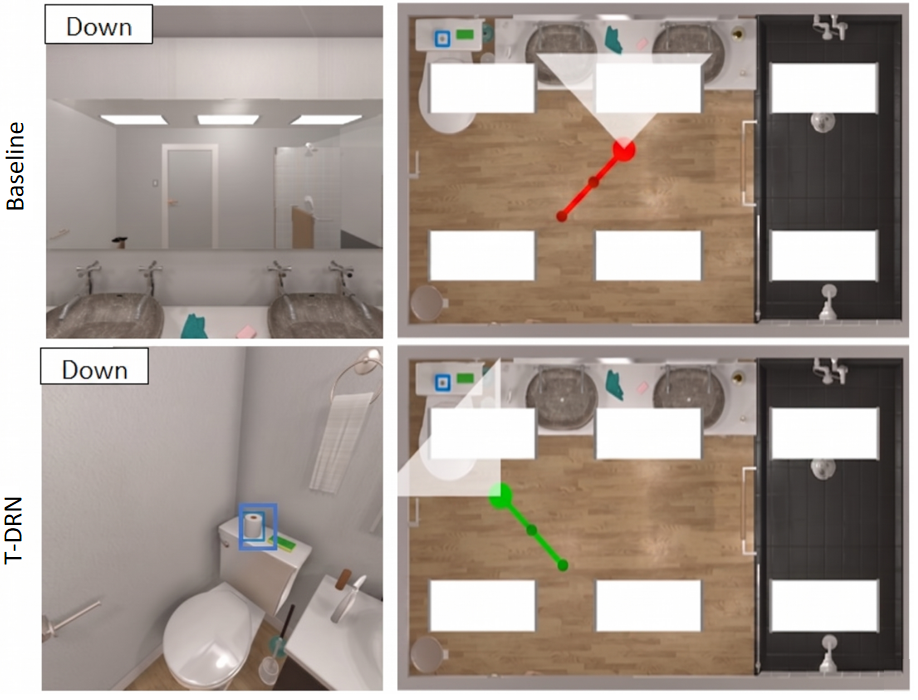}
    \caption{Qualitative comparison between TDANet and the proposed T-DRN in a zero-shot scenario. The target object (toilet paper) is unseen during training. TDANet is distracted by visually similar objects, while T-DRN successfully navigates by leveraging relational differences among all detected objects.}
    \label{compare}
\end{figure}

Fig.~\ref{compare} illustrates a zero-shot navigation example where the target object (toilet paper) is unseen during training. TDANet is distracted by a towel due to high attention weight, resulting in incorrect navigation. In contrast, T-DRN evaluates relational differences across all detected objects, enabling the agent to identify the correct target and reach the goal successfully.

\subsection{Model Efficiency and Qualitative Analysis}
\begin{table}[h]
\centering
\caption{Comparison of parameter count and inference time across different models.}
\label{tab:efficiency}
\begin{tabular}{@{}lcc@{}}
\toprule
\textbf{Model} & \textbf{Params (M)}  & \textbf{Latency (ms)}  \\ \midrule
MJOLNIR-o \cite{mjol} & 3.42 & 0.6 \\
SSNet \cite{SSNet} & 2.00 & 0.5 \\
TDANet \cite{tdanet} & 3.08 & 0.6 \\ \midrule
\textbf{T-DRN (Ours)} & \textbf{2.96} & \textbf{0.5} \\ \bottomrule
\end{tabular}
\end{table}

For deployment on resource-constrained mobile robots, maintaining low computational overhead is crucial. Table~\ref{tab:efficiency} compares the model complexity and inference latency of T-DRN against several representative models. All inference times are measured on a single NVIDIA GeForce RTX 3060 GPU to ensure fair comparison. As shown in Table~\ref{tab:efficiency}, T-DRN contains only 2.96M parameters, making it more lightweight than MJOLNIR-o (3.42M) and TDANet (3.08M). Furthermore, T-DRN achieves an inference latency of 0.5\,ms per frame, matching the fastest model (SSNet) while delivering superior navigation performance. These results demonstrate that T-DRN provides an effective balance between performance and computational efficiency, making it well-suited for real-time deployment on resource-constrained robotic platforms.

\subsection{Real-World Deployment} 

To evaluate sim-to-real transfer directly, we deployed the policy trained in simulation onto a real-world robot TurtleBot4 without additional fine-tuning. YOLOv7\cite{yolov7} was used to extract bounding boxes and their corresponding category labels from robot's RGB observations. Communication with the robot was implemented through ROS2. We evaluated the system in three real-world settings---office, bedroom, and living room---for a total of 39 trials.

\begin{table}[htbp]
\centering
\caption{Real-world zero-shot navigation performance on a TurtleBot4 platform. T-DRN achieves the highest SR, demonstrating superior sim-to-real transfer capability.}
\renewcommand{\arraystretch}{1.2}
\begin{tabular}{@{}lc@{}}
\toprule
Model & SR (\%) \\
\midrule
MJOLNIR \cite{mjol} & 20.5 \\
TDANet \cite{tdanet} & 53.8 \\
\textbf{T-DRN (ours)} & \textbf{64.1} \\
\bottomrule
\end{tabular}
\label{tab:physical_results}
\end{table}
\begin{figure*}[t]  
    \centering
    \includegraphics[width=\textwidth]{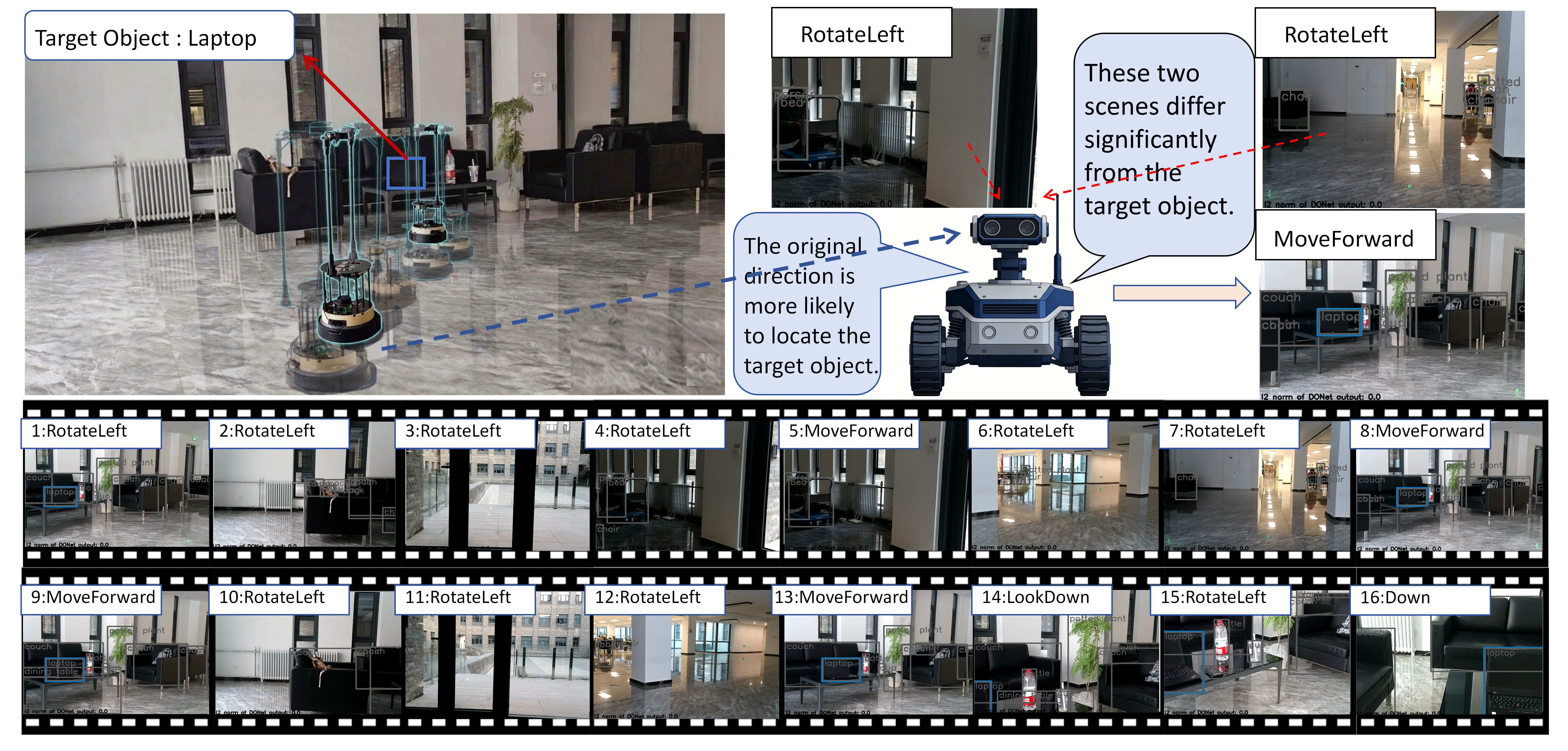}  
    \caption{Real-world zero-shot object-goal navigation using T-DRN. The top panel presents the agent trajectory and decision process. The bottom filmstrip shows the sequential egocentric observations and corresponding actions. The agent evaluates candidate directions using relational differences and maintains short-term memory to handle narrow FoV conditions.}
    \label{fig:model}
\end{figure*}

Table~\ref{tab:physical_results} presents the real-world experimental results. T-DRN achieves the highest SR and improves over TDANet by 10.3\%, indicating that the proposed representation transfers more realiably under real sensing constraints.

Figure~\ref{fig:model} illustrates a real-world zero-shot object-goal navigation example. The top panels demonstrate how T-DRN leverages difference-based relational learning and a dual-frame temporal buffer to achieve stable navigation behavior.

The agent assess candidate orientations based on their relational difference with respect to the target object. As illustrated by the ``significant difference'' bubble, the difference-based feature extractor measures the discriminative gap between the current scene (e.g., warm radiators or glass corridors) and the target. When a candidate scene deviates excessively, the agent rejects that direction (indicated by 'RotateLeft' actions) and continues along a more promising search trajectory.

Crucially, this persistency is enabled by the dual-frame temporal buffer, highlighted by the bubble stating ``the original direction is more likely.'' Due to the robot's restricted FoV, the agent frequently loses visual contact with potential targets during rotations (e.g., between panels 5 and 6, or 12 and 13 in the bottom filmstrip). By maintaining a dual-frame temporal memory, T-DRN preserves a form of object permanence. This allows the agent to reject a noisy, high-disparity candidate view and persist in a promising search direction, thereby compensating for information loss caused by the constrained sensing window.

In contrast, MJOLNIR \cite{mjol} achieved a much lower SR in the physical experiments because it relies on a predefined object-relation gragh to model hierarchical relationships. This graph is limited to the 22 major object categories used in the simulation data, whereas many real-world objects lack the semantic information required by the model. This mismatch subsequently degrades MJOLNIR's real-world performance. TDANet, while much stronger, still frequently loses candidate objects because of the restricted camera FoV and its dependence on instantaneous attention allocation. By contrast, our model partially mitigates these failures through explicit short-term temporal memory.

A primary failure mode of T-DRN arises from its RGB-only nature. Lacking explicit depth perception, the agent occasionally struggles to accurately estimate the spatial proximity of obstacles, leading to collision-induced navigation failures.

\section{Discussion}
\subsection{Ablation Study}
To evaluate the individual contributions of the core components in the proposed T-DRN, we conducted an ablation study in the real-world environment. We systematically isolated two primary modules: the Difference-Based Relational Learning and the Dual-Frame Temporal Buffer. The baseline model utilizes a standard single-frame object encoder without explicit relational difference calculations. The quantitative results are summarized in Table \ref{tab:ablation}. Both components contribute substantially to the overall success of the zero-shot navigation policy.
\begin{table}[h]
\centering
\caption{Ablation study of T-DRN components in real-world zero-shot navigation. }
\label{tab:ablation}
\begin{tabular}{@{}lcc|c@{}}
\toprule
\makecell{Model \\ Variant} & \makecell{\textbf{Diff} \\ \textbf{Module}} & \makecell{\textbf{Temp} \\ \textbf{Module}} & SR(\%) \\ \midrule
Baseline & $\times$ & $\times$ &  7.7 \\
Difference Only & $\checkmark$ & $\times$ & 51.3 \\
Temporal Only & $\times$ & $\checkmark$ & 18.0 \\ 
\textbf{T-DRN (Ours)} & $\checkmark$ & $\checkmark$ & \textbf{64.1} \\ \bottomrule
\end{tabular}
\end{table}

\textbf{Effectiveness of Difference-Based Learning:} 
Comparing the baseline with the Difference-Only variant reveals that explicitly computing relational difference  improves SR by 43.6\%. This result supports our hypothesis that relational features are less sensitive to domain-specific appearance bias than absolute visual recognition, and therefore generalize better to unseen target categories.

\textbf{Effectiveness of Temporal Buffer:} 
Integration the dual-frame temporal memory module leads to an improvement of 10.3\% in SR. This gain is consistent with improved short-term spatial continuity. By maintaining object permanence over two frames, the temporal buffer reduces policy collapse and redundant rotational actions when candidate objects temporarily leave the restricted sensing window. The full T-DRN framework, which combines both modules, yields the highest performance, indicating that relational-disparity features are most effective when temporally grounded.

\subsection{Temporal Memory Analysis}
\label{subsec:design_params}

To investigate the impact of temporal memory span on navigation performance, we evaluate configurations across various time horizons. Furthermore, we introduce a learnable attention mechanism to adaptively fuse historical features for comparative analysis. The procedure for computing attention weights over $T$ steps is outlined in Algorithm~\ref{alg:temporal_attention}.

\begin{table}[htbp]
\centering
\caption{Performance comparison across different temporal memory lengths. The proposed dual-frame temporal buffer (2-step history) achieves the best performance.}
\label{tab:sr_comparison}
\begin{tabular}{lcc}
\toprule
\multirow{2}{*}{\textbf{Temporal History Length}} & \multicolumn{2}{c}{\textbf{SR (\%)}} \\
\cmidrule(lr){2-3}
 & $L\geq1$ & $L\geq5$ \\
\midrule
10-step history & 54.1 & 33.4 \\
5-step history & 61.2 & 45.8 \\
3-step history & 65.4 & 46.2 \\
5-step + attention & 62.7 & 42.7 \\
\textbf{2-step (Ours)} & \textbf{71.9} & \textbf{57.2} \\
\bottomrule
\end{tabular}
\vspace{0.1cm}

\end{table}

The experimental results are presented in  Table~\ref{tab:sr_comparison}. The dual-frame temporal memory model achieves the overall best performance among the tested memory configurations while also remaining   computationally simpler than attention-based alternatives. Interestingly, shorter temporal history improves performance, suggesting that long-term memory may introduce noise under restricted FoV conditions.
\begin{algorithm}[t]
\caption{Learnable Temporal Attention}
\label{alg:temporal_attention}
\begin{algorithmic}[1] 
\Require Historical features $\mathbf{H} =[\mathbf{h}_1,\ldots,\mathbf{h}_T]^\top \in \mathbb{R}^{T \times D}$
\Ensure Attention weights $\mathbf{w} \in \mathbb{R}^T$, weighted features $\tilde{\mathbf{H}} =[\tilde{\mathbf{h}}_1,\ldots,\tilde{\mathbf{h}}_T]^\top \in \mathbb{R}^{T \times D}$

\State \textbf{Initialize:} Temporal bias vector $\boldsymbol{\beta} \in \mathbb{R}^T$ ($\beta_1 < \dots < \beta_T$)
\State $\bar{\mathbf{h}} \gets \frac{1}{T}\sum_{i=1}^T \mathbf{h}_i$
\State $g \gets \sigma(\mathbf{w}_g^\top \bar{\mathbf{h}} + b_g)$ \Comment{Dynamic gating scalar}

\State $\mathbf{s}^c \gets \mathbf{H} \mathbf{w}_c + b_c$ \Comment{Content-based scores}
\State $\mathbf{s} \gets (1-g)\mathbf{s}^c + g\boldsymbol{\beta}$ \Comment{Score fusion}

\State $\mathbf{w} \gets \text{softmax}(\mathbf{s})$

\State $\tilde{\mathbf{h}}_i \gets w_i \mathbf{h}_i, \quad \forall i \in \{1, \dots, T\}$ \Comment{Apply attention weights}

\State \textbf{return} $\tilde{\mathbf{H}}, \mathbf{w}$
\end{algorithmic}
\end{algorithm}

\subsection{Object Aggregation Analysis}
To further investigate the effectiveness of our feature aggregation mechanism, we conducted a parametric study on the number of objects involved in the scene representation. In the proposed T-DRN, the scene-level representation $X_t$ is generated by averaging the difference vectors of \textbf{all} detected objects in the FoV. We compared this strategy against a Top-$k$ selection approach, where only the $k$ objects with the smallest discriminative gap (i.e., those most similar to the target) were selected for aggregation.

We evaluated the performance across different values of $k \in \{2, 3, 5, 7\}$. The experimental results, summarized in Table~\ref{tab:topk_experiment}, demonstrate that the all-object aggregation strategy consistently outperforms the Top-$k$ selection across the reported metrics. We conjecture this phenomenon is attributed to two main factors.
1) \textbf{Contextual Navigation Cues}: While the Top-$k$ objects are semantically closer to the target, the excluded objects often serve as vital spatial anchors or environmental cues that help the agent understand its relative position and orientation within the scene.
2) \textbf{Information Robustness}: Under restricted FoV conditions, the number of detected objects is already limited. Further filtering the objects through Top-$k$ selection intensifies information loss, leading to unstable policy outputs and decreased navigation success rates.

\begin{table}[h]
\centering
\caption{Comparison of object aggregation strategies. Aggregating all detected objects achieves the best performance.}
\label{tab:topk_experiment}
\begin{tabular}{@{}lccc@{}}
\toprule
Strategy & $k$ & SR (\%)  & SPL (\%)   \\ \midrule
Top-$k$ selection & 2   & 22.8 & 6.3 \\
Top-$k$ selection & 3   & 37.2 & 7.4 \\
Top-$k$ selection & 5   & 36.0 & 10.1 \\
Top-$k$ selection & 7   & 39.2 & 11.1 \\ 
\textbf{All Objects (Ours)} & - & \textbf{62.7} & \textbf{20.2} \\ \bottomrule
\end{tabular}
\end{table}
The results indicate that narrowing the focus to only a few ``target-like'' objects through Top-$k$ selection negatively impacts the agent's spatial awareness. In zero-shot scenarios, the agent's ability to navigate depends not only on identifying the goal but also on perceiving the relative layout of the environment. By aggregating features from all observed objects, our model preserves these implicit environmental anchors, which proves essential for maintaining a stable trajectory when the target is not yet within the immediate line of sight.

\section{Conclusion}
In this letter, we proposed T-DRN, a lightweight end-to-end framework for zero-shot object-goal navigation. By shifting the learning paradigm from absolute feature recognition to difference-based relational modeling, the proposed method demonstrates superior generalization to unseen target categories.
To bridge the gap between simulation and real-world deployment, we introduced a dual-frame temporal memory mechanism. This temporal buffering strategy mitigates the performance degradation caused by geometric discrepancies, particularly the restricted FoV of physical robotic platforms. Experimental results in AI2-THOR  and real-world deployments on a Turtlebot4 confirm that T-DRN outperforms the compared baselines in both success rate and navigation efficiency. Although the physical evaluation is not intended to be an exhaustive benchmark, it demonstrates the practical feasibility of the proposed representation under real sensing and control constraints. These results suggest that temporal consistency and global scene context are important for transferring learned navigation policies from simulation to constrained real-world scenarios.

Future work will explore the integration of long-term memory structures to handle more complex, multi-room environments and further investigate the impact of dynamic obstacles on difference-based feature stability.



\bibliographystyle{IEEEtran}
\bibliography{IEEEabrv,mybibfile}

\end{document}